\documentclass[sigconf]{acmart}
\AtBeginDocument{%
  \providecommand\BibTeX{{%
    \normalfont B\kern-0.5em{\scshape i\kern-0.25em b}\kern-0.8em\TeX}}}

\copyrightyear{2024} 
\acmYear{2024} 
\setcopyright{acmlicensed}\acmConference[HRI '24 Companion]{Companion of the 2024 ACM/IEEE International Conference on Human-Robot Interaction}{March 11--14, 2024}{Boulder, CO, USA}
\acmBooktitle{Companion of the 2024 ACM/IEEE International Conference on Human-Robot Interaction (HRI '24 Companion), March 11--14, 2024, Boulder, CO, USA}
\acmDOI{10.1145/3610978.3640560}
\acmISBN{979-8-4007-0323-2/24/03}

\begin{document}

\title{Adversarial Robots as Creative Collaborators}

\author{Shayla Lee}
\email{shayladlee@gmail.com}
\orcid{0009-0005-9552-2454}
\affiliation{%
  \institution{The New School \& Cornell Tech}
  \city{New York}
  \state{NY}
  \country{USA}
}

\author{Wendy Ju}
\email{wendyju@cornell.edu}
\affiliation{%
  \institution{Cornell Tech}
  \city{New York}
  \state{NY}
  \country{USA}
}

\renewcommand{\shortauthors}{Shayla Lee \& Wendy Ju}

\begin{abstract}
  This research explores whether the interaction between adversarial robots and creative practitioners can push artists to rethink their initial ideas. It also explores how working with these robots may influence artists’ views of machines designed for creative tasks or collaboration. Many existing robots developed for creativity and the arts focus on complementing creative practices, but what if robots challenged ideas instead? To begin investigating this, I designed UnsTable, a robot drawing desk that moves the paper while participants (\textit{N}=19) draw to interfere with the process. This inquiry invites further research into adversarial robots designed to challenge creative practitioners.
\end{abstract}






\maketitle

\section{Introduction}
There have been many robots developed for creativity and art. Despite these robots coming to fruition, traditional artists have not yet incorporated them into their creative processes. Many of these machines, like 3D printers, for example, are designed to make things, and this has been historically intimidating to some artists \cite{devendorf2016strange}. Tensions between artists and technology have been demonstrated through the arts and crafts movement \cite{Open_University_2013}, the reluctance to accept photography as an art form \cite{Teicher_2016}, and more recently through the discussions around Generative Artificial Intelligence.

It is true that some creatives, like Sougwen Chung \cite{Chung}, have embraced technology fully, but there is still a gap between the artist and the machine. Robots focused on art and creativity tend to mimic, complement, and generate content from what already exists \cite{schaldenbrand2022frida, 10.1145/3461778.3462116}. These robotic tools and experiments don’t address the value of conflict in the early stages of the creative process.

When designing or creating something, it is productive to have an influence that pushes back on ideas, suggests alternatives, or incites reactions to unexpected interventions throughout the process \cite{devendorf2016strange}. In addition to running ideas by other people \cite{ju2006thinking}, many artists will do challenges or play games to spark their creativity or get unstuck. Some professional creative practitioners have designed products for this purpose, like the “Oblique Strategies” cards, by Brian Eno and Peter Schmidt. Each card makes an unexpected suggestion to inspire new tracks of thinking and break through creative blocks \cite{Eno_2001}. An adversary is someone or something that works against you, and this role seems to be unfilled by the creativity-oriented machines that have been produced thus far. 
\begin{figure}[t]
  \centering
  \includegraphics[width=\linewidth, trim={0 10cm 0 5cm},clip]{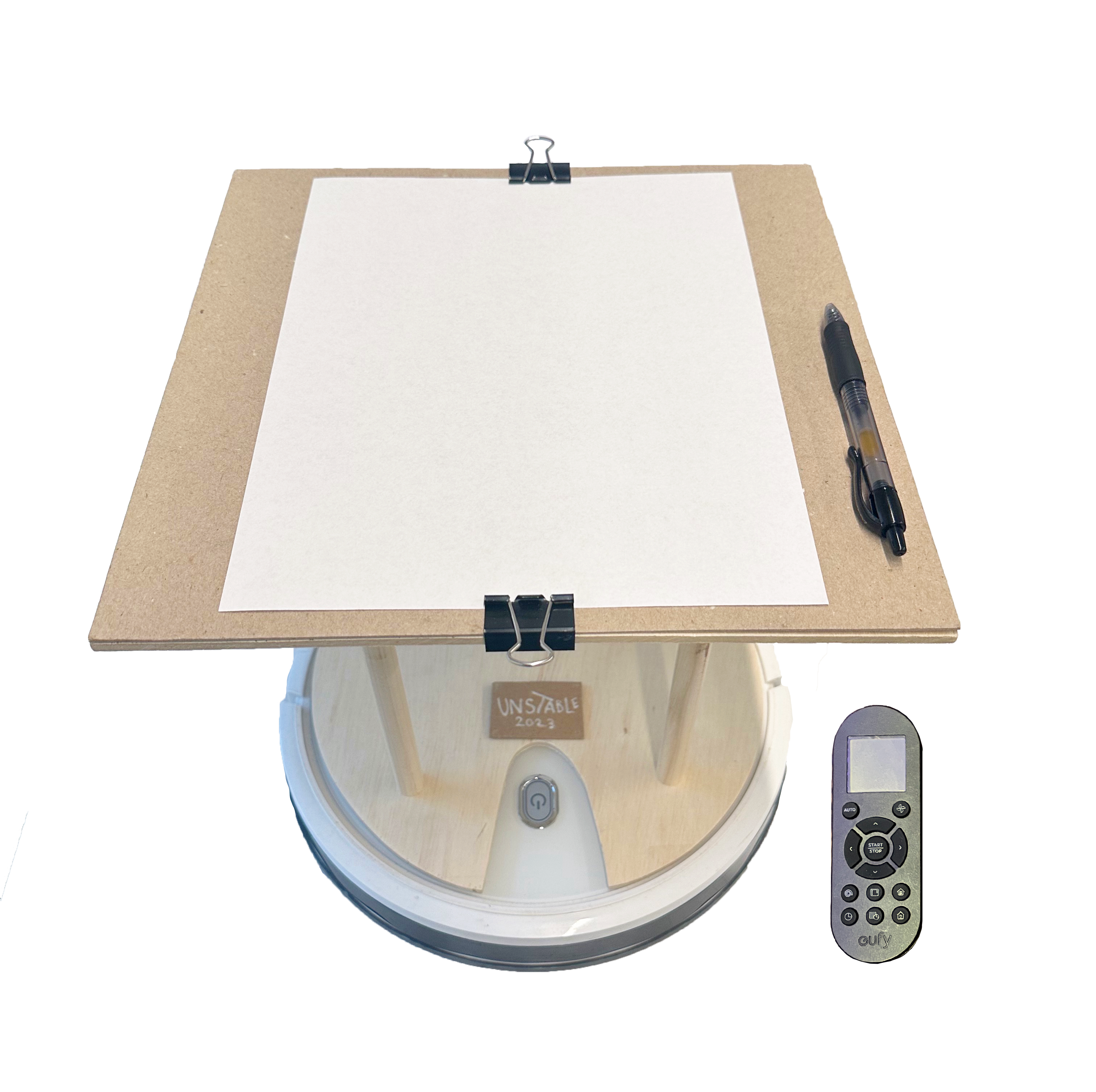}
  \caption{UnsTable is a remote-operable Eufy vacuum robot with a table attachment that moves while a participant is drawing on it. The wooden table is topped with chipboard and holds a piece of paper for the drawing.}
  \Description{A robot vacuum with a square wooden desk attached to the top. There is a paper clipped to the top of the desk area, a pen beside the paper, and a remote in the bottom right corner of the image.}
  \label{fig1}
\end{figure}

As an initial exploration into collaborative adversarial robots for artists, I designed UnsTable, a robot drawing desk that moves a paper while a participant draws to interfere with the process. The current state of this research is not intended to present conclusive evidence for the effectiveness of UnsTable as an adversarial robot. It is meant to share findings from experiments containing adversarial interactions and present this inquiry for further exploration. 

\section{Methods}

\begin{figure}
  \centering
  \includegraphics[width=\linewidth, trim={5cm 10cm 10cm 0},clip]{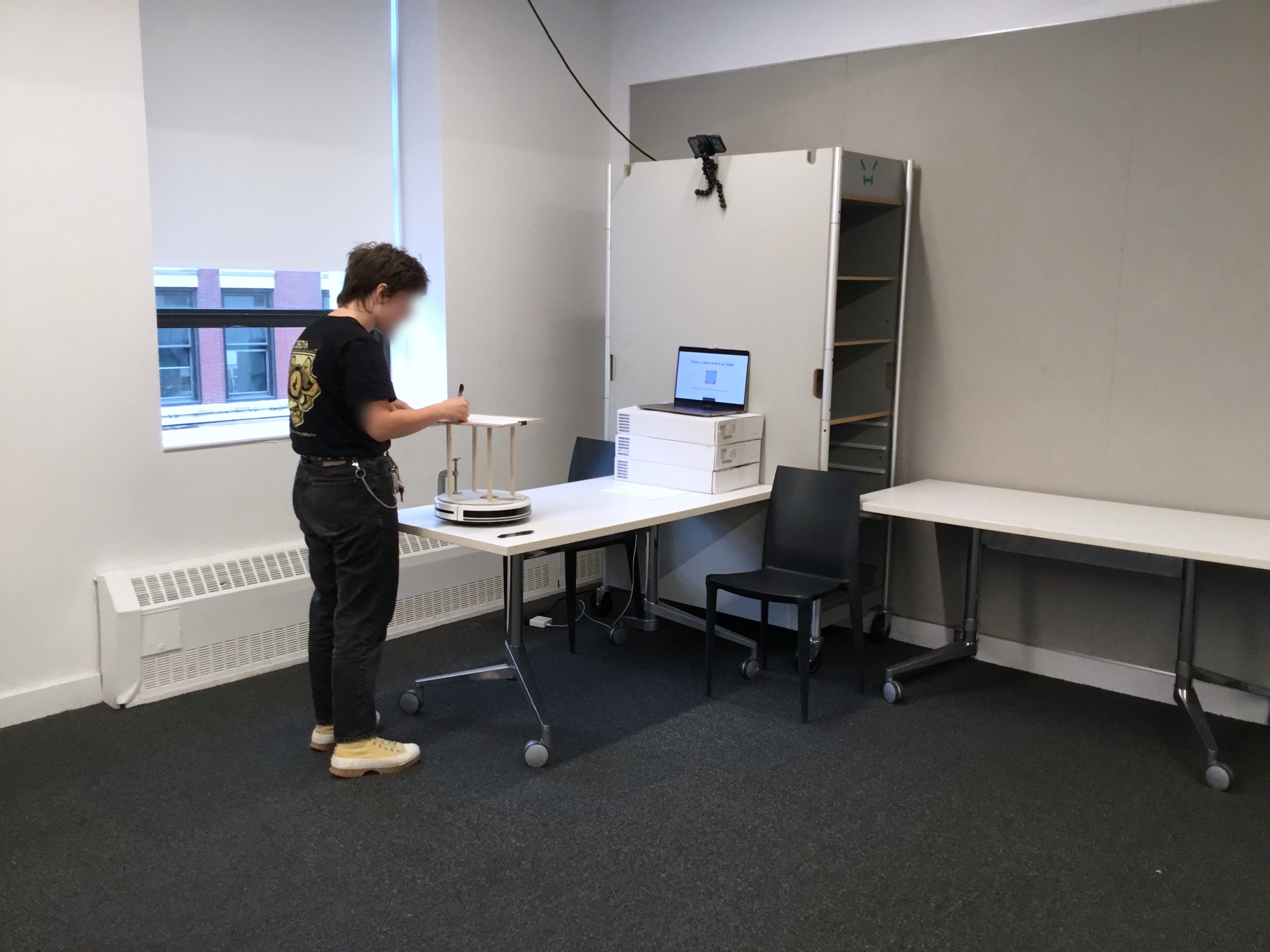}
  \caption{Pilot study participant stands, ready to draw with a pen, in front of UnsTable on table. Note instrumentation.}
  \Description{A zoomed-out photo of the pilot study setup that shows the participant, UnsTable on a long desk turned sideways in front of them, the computer with a prompt at the far end of the table, and a phone camera above the computer to record the drawing process. UnsTable extends the desk surface to the participant's elbow level.}
  \label{fig2}
\end{figure}
UnsTable (Figure \ref{fig1}) is a Eufy robot vacuum with a platform attached to the top, intended to be placed on a table to transform it into a standing desk. The Eufy is remote-operable, so all movement is currently performed using the Wizard of Oz Method \cite{fakeitWoZ}. UnsTable can be moved forward or backwards and rotate left and right when the buttons are pressed on the remote control. It also has built-in infrared edge detection which prevents it from falling off the table.

\subsection{Platform Testing}
Initial informal testing of the UnsTable platform engaged \emph{N}=12 participants (primarily undergraduate students and faculty) in a pop-up exhibition as well as an undergraduate research presentation for design students. 
During platform testing, participants were asked to use UnsTable to draw. Participants could choose to use pencils or charcoal, had the option to erase, and could make as many drawings as they desired. As they drew, UnsTable would move the paper and drawing surface. UnsTable could occasionally spin in a circle for several seconds, for example.

Participants were not informed that the robot was being moved using a remote control, although some who participated later on in each session had already found out that it was being controlled by a person. UnsTable's movements were improvised by the remote operator based on preference and experimentation rather than a predetermined plan for movement as was used in the subsequent pilot study.

\subsection{Pilot Study}
The following pilot study was designed with more consistency in setup, documentation, delivery, and materials, as shown in Figure \ref{fig2}. It was conducted at an undergraduate student project fair, and was tested by \emph{N}=7 participants (six undergraduate students and one faculty member). The setup included two cameras: one iPhone camera mounted several feet above the table to record the drawing and the movement of UnsTable from above, and a Zoom call hosted on the laptop in front of participants to record their reactions and facial expressions during each interaction. The laptop screen also displayed the participant’s prompt and included a QR code to a follow-up survey.

Each participant was given one of two prompts, beginning with “This is UnsTable, it is a drawing desk that moves while you're working to make suggestions to what you’re drawing,” and followed either by “Please draw a robot” or “You can draw anything you like.” Of the seven participants, six received the robot prompt (Figure \ref{fig4}) and one received the free-draw prompt (Figure \ref{fig3}, Drawing I). Due to time restraints and the quantity of un-prompted drawings produced during platform testing, this ratio was chosen to collect data from participants using UnsTable with more restrictive conditions where they are given a specific task. All participants used the same pen (Pilot G-2 07), began with a portrait-oriented paper which was secured to the drawing surface with binder clips, and only produced one drawing each.

For the pilot study, the remote operator drove UnsTable according to the following algorithm: “If a person keeps drawing in one area for three seconds OR they start a new line, then press two movement buttons. Possible combinations include a mix of forward or backward and turning, or can be two consecutive movements of the same type (for example forward + forward or right + right).” This set of directives was selected for their performability. Once the study began, UnsTable was moved twice on a three-second interval, rather than triggering movement based on participant behavior, since the intended conditions were difficult for the operator to observe and carry out reliably and effectively.

All seven participants completed a follow-up survey after they finished their drawing. The survey asked if the participant consider themselves to be an artist, if they draw as part of their practice, if they started with an initial idea for the drawing, if it changed while working with UnsTable, their level of enjoyment, level of frustration, whether or not they wanted to stop using UnsTable at any point, and what they believed UnsTable’s role is (with the options of “Collaborator”, “Assistant”, “Tool”, and “Obstacle”).

\section{Results}
    \subsection{Platform Testing}
    \begin{figure*}
      \centering
      \includegraphics[width=\linewidth]{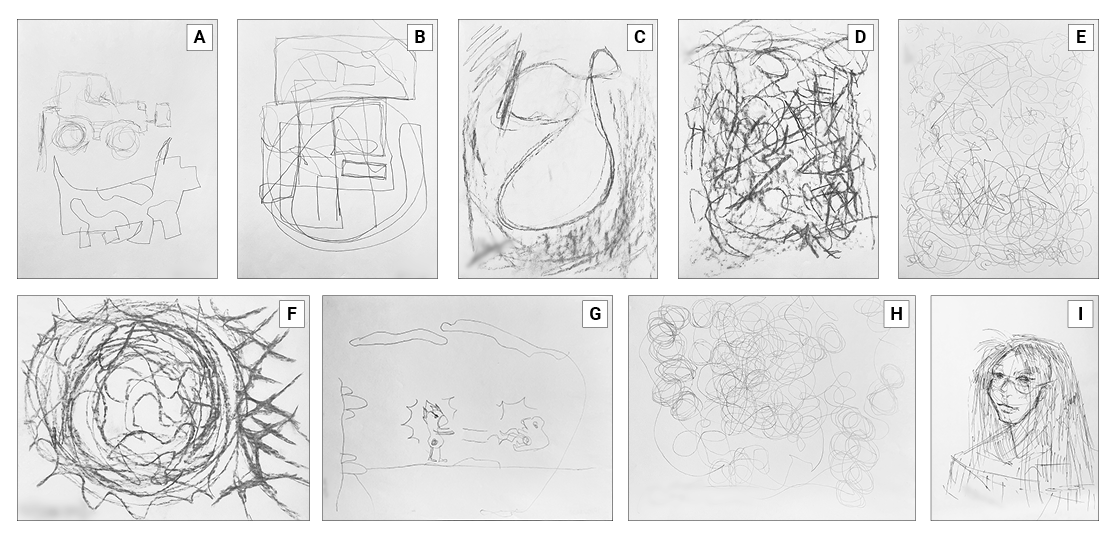}
      \caption{A sample of drawings using various mediums (charcoal, pencil, pen) by eight participants from the platform test and pilot study who received an open prompt. Their drawings include abstract and objective subjects as well as other doodle experiments.}
      \Description{Nine black and white drawings feature a mix of scribbles and representational sketches. Six of the drawings are portrait oriented while the remaining three are landscape.}
      \label{fig3}
    \end{figure*}
The first set of platform tests with UnsTable resulted in 17 drawings. Since no quantitative data was collected during the course of these tests, we have compiled some qualitative observations we believe are significant to the study.

It was observed that participants who made more than one drawing tended to lean into the process further after their first attempt, especially if they had aimed for something more deterministic in the first drawing. In these cases, they also verbally reported being less frustrated the second time. This is demonstrated in Figure \ref{fig3}, where Drawings A and B were made by the same person in the order they appear in the figure. The first drawing includes a sketchy tractor and an oddly shaped cow while the second is a series of looser, darker, abstract shapes and lines which occupy most of the page.  

One student elected to use a reference image of the number seven to guide their drawing. As UnsTable moved, it instigated the addition of novel artifacts like a loop in the corner of the number's outline as well as a wide, curvy stem (Figure \ref{fig3}, Drawing C), despite this not matching the original image. This indicated an openness to collaboration despite UnsTable interfering with their original vision.

Another student changed the orientation of their drawing from portrait to landscape mid-way through because of the way the paper faced them after the robot had finished a turn (Figure \ref{fig3}, Drawing G). The same participant also incorporated any new marks created with the robot into a story about two battling birds that guided the rest of the drawing. 

While some participants were fine artists or experienced illustrators, others were not typically drawn to sketching. One student who reported having minimal drawing experience ended up making two pages full of stars and spirals (Figure \ref{fig3}, Drawings D and E). They said that UnsTable had made them want to draw with it because they recognized, after watching other participants, that the output was never perfect and didn't have to be. Throughout the platform testing, there were consistently several bystanders who watched each interaction and seemed excited to get to try UnsTable for themselves. 

Later in the platform testing, onlookers and participants had begun collectively experimenting with UnsTable. One student, for example, let the robot draw by holding the pencil still, allowing the way it moved the paper to determine which marks were made. A faculty member then, similarly, drew the same shape over and over again with a pencil and let the robot make most of the decisions for the drawing (Figure \ref{fig3}, Drawing H). Although they relinquished control of the result, they were delighted each time an interesting move was made and consistently had smiles on their faces. 

Participants generally reported enjoying the process of working with UnsTable, asked many questions, and one participant even attended two separate platform testing sessions to use UnsTable a second time (Figure \ref{fig3}, Drawing F).

\subsection{Pilot Study}
    \begin{figure}
      \centering
      \includegraphics[width=\linewidth]{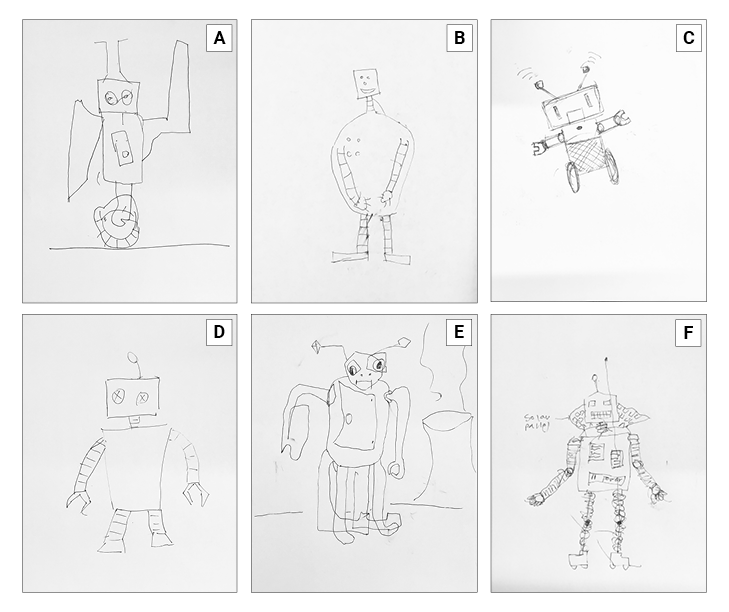}
      \caption{Six participants were asked to draw a robot as part of the pilot study, and they reported adding details and changing aspects of their drawings they wouldn’t have without UnsTable’s intervention.}
      \Description{Six robot drawings in black pen on white paper. Each one has a head with two eyes, a torso with two arms, some have legs and some have wheels. They feature odd lines, details, and extra strokes from using UnsTable as a drawing surface.}
      \label{fig4}
    \end{figure}
Each of the seven participants in the pilot study made one drawing. Unlike the initial platform tests, all seven participants maintained a portrait-oriented drawing. Both frustration and enjoyment were observed during the tests (and this is also reflected below in the survey results). 

In the survey, all participants said that they consider themselves artists though only three of them said they specifically draw or sketch often. Most (five out of seven) participants reported that they changed their drawings at least a little bit while working. Some of the explanations for how the drawings changed included making a robot with a round body rather than one that's square (Figure \ref{fig4}, Drawing B), changing the position of the head in a portrait (Figure \ref{fig3}, Drawing I), and adjusting their drawing method by using more straight lines than curves to maintain control of the output (Figure \ref{fig4}, Drawing C). Two participant more generally summarized that the drawing changed multiple times, "Whenever UnsTable changes its direction or orientation, the drawing changes how I wanted [it] to be", or that they "Rolled with some of the things that happened when [they were] drawing (bigger eyes, weirdly shaped hands, etc.)" (Figure \ref{fig4}, Drawing E). 

The level of enjoyment and frustration were rated on a five-point scale. The results of the survey indicated that participant experiences were typically enjoyable (M=4.3, SD=0.5) and that they experienced moderate frustration (M=2.57, SD=1.0). Participants described their experiences as "extremely funny", challenging, "open ended and free", and "a very novel experience" in response to why they enjoyed working with it. One participant expressed having fun but wished their drawing had turned out better, while another said it had made them focus and be more careful while simultaneously trying to use the robot's movements to intentionally add to their drawing. One participant reported that it was frustrating to not achieve their intended result, and another agreed but said the "weird lines... became one part of the art". Participants were also asked whether they had wanted to stop working with UnsTable at any point in the process, and two said yes due to frustration.

“Collaborator” was the most commonly selected role, with three out of seven choosing it in the survey. The remaining four participants were split evenly between “Tool” and “Obstacle”. No participants selected “Assistant” as the best-fit description for UnsTable. 

\section{Discussion}
Based on the smiles and laughter of participants among other observations, our impression from both the platform testing and pilot study is that adversarial interactions with UnsTable resulted in enjoyment, curiosity, and experimentation. It’s interesting, for example, that some users allowed UnsTable to lead the creative decision-making by acting robotic and limiting their own movement. This may suggest a shift in perception of the robot, which was one of the considerations in the scope of this research. Additionally, many of the participants reported changing their initial idea, alluding to the instigation of new creative explorations through these interactions. 

There are many limitations to these preliminary studies, as they are inherently incomplete. 
The participant sample size was small, and primarily involved undergraduate design students. Expanding the user testing and analysis to settings beyond the design school setting would likely provide richer more nuanced insights into the perceptions, challenges and creative opportunities presented by adversarial robot interventions. 

Some participants were also aware that UnsTable was being remote-controlled, and this could have skewed responses to it. Further iterations of this research would likely involve an automated UnsTable, and feature more movement patterns. Though the construction of UnsTable may be simple, we acknowledge the exact conditions of this preliminary study may be difficult to replicate. It is also worth acknowledging potential cultural biases in the conception of the guiding questions of these preliminary inquiries. 

As aforementioned, this research is an invitation to continue the investigation of adversarial interactions between robots and creative practitioners. Perhaps adversarial interactions can bring artists and technology together and change the future of robots as creative collaborators.

\begin{acks}
We want to thank Justin Bakse who encouraged Shayla to share her research with the world. His time, feedback, connections, and support made all the difference. The NSF Research Experiences for Undergraduates program supported Shayla's introduction to the Future Automation Research Lab at Cornell Tech. Thanks to Maria Teresa Parreira, David Goedicke, Natalie Friedman, Dave Dey, and others from the lab, and to Shayla's Parsons Design and Technology thesis faculty Melanie Crean, Ernesto Klar, and Alexander King. Finally, we want to also thank Harpreet Sareen, Kyle Li, Anthony Dunne, Jim McCann, Carla Diana, Sven Travis, Margaret Rhee, Mattia Casalegno, Arielle Mella, Jacob Hennessy-Rubin and those in the NY Robotics Network, whose perspectives on art, creativity, and robotics helped shape this research.

\end{acks}

\bibliographystyle{ACM-Reference-Format}
\bibliography{Main.bib}


\end{document}